# Losing Focus: Can It Be Useful in Robotic Laser Surgery?

N.E. Pacheco, Y.A. Garje, A. Rohra, L. Fichera

*Department of Robotics Engineering*
*Worcester Polytechnic Institute, USA*
nepacheco@wpi.edu

## INTRODUCTION

Lasers are an essential tool in modern medical practice, and their applications span a wide spectrum of specialties. In laryngeal microsurgery, lasers are frequently used to excise tumors from the vocal folds [1]. Several research groups have recently developed robotic systems for these procedures [2-4], with the goal of providing enhanced laser aiming and cutting precision.

Within this area of research, one of the problems that has received considerable attention is the automatic control of the laser focus. Briefly, *laser focusing* refers to the process of optically adjusting a laser beam so that it is concentrated in a small, well-defined spot – see Fig. 1. In surgical applications, tight laser focusing is desirable to maximize cutting efficiency and precision; yet, focusing can be hard to perform manually, as even slight variations (< 1 mm) in the focal distance can significantly affect the spot size. Motivated by these challenges, Kundrat and Schoob [3] recently introduced a technique to robotically maintain constant focal distance, thus enabling accurate, consistent cutting. In another study, Geraldes et al. [4] developed an automatic focus control system based on a miniaturized varifocal mirror, and they obtained spot sizes as small as 380 μm for a $CO_2$ laser beam.

Whereas previous work has mainly dealt with the problem of creating – and maintaining – small laser spots, in this paper we propose to study the utility of defocusing surgical lasers. In clinical practice, physicians defocus a laser beam whenever they wish to change its effect from cutting to heating – e.g., to thermally seal a blood vessel [5]. To the best of our knowledge, no previous work has studied the problem of robotically regulating the laser focus to achieve controlled tissue heating, which is precisely the contribution of the present manuscript.

In the following sections, we first briefly review the dynamics of thermal laser-tissue interactions and then propose a controller capable of heating tissue according to a prescribed temperature profile. Laser-tissue interactions are generally considered hard to control due to the inherent inhomogeneity of biological tissue [6], which can create significant variability in its thermal response to laser irradiation. In this paper, we use methods from nonlinear control theory to synthesize a temperature controller capable of working on virtually any tissue type without any prior knowledge of its physical properties.

## MATERIALS AND METHODS

**Problem Formulation**. Let us consider a scenario where a tissue specimen is exposed to a laser beam of intensity

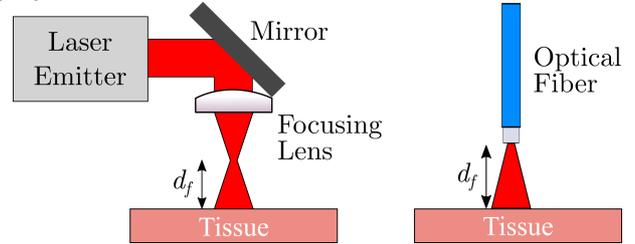

**Fig. 1:** The goal of laser focusing is to create a spot size of prescribed width via the control of $d_f$, i.e., the distance between the laser beam's focal point and the tissue surface. (Left) In *free beam* systems, the location of the focal point depends on the characteristics of the lenses used to focus the beam. (Right) In fiber-based systems, laser light diverges immediately upon exiting the fiber, with an angle determined by the numerical aperture of the fiber itself. In this manuscript, we study how regulating $d_f$ can be used to produce controlled tissue heating.

$I$ (W/cm$^2$). The problem we wish to solve is to control the tissue temperature at the point of incidence of the laser. We assume that the laser beam can only be moved vertically with respect to the tissue, i.e., that the only variable we can control is the distance $d_f$ between the tissue surface and the beam's focal point – refer to Fig. 1.

**Preliminaries.** From [7], the temperature dynamics of laser-irradiated tissue can be modeled as:

$$c_v \frac{\partial T}{\partial t} = \kappa \nabla^2 T + \mu_a I \quad (1)$$

where $T$ represents the tissue temperature as a function of space and time, and $c_v$, $\kappa$, and $\mu_a$ are three tissue-specific physical parameters – namely, the *volumetric heat capacity*, the *thermal conductivity*, and the *coefficient of absorption*. We note that these parameters are rarely known with certainty, as different types of tissue will generally have different physical properties, and significant variations are possible even within specimens of the same tissue type [6].

We can regulate the beam intensity $I$ in Eq. (1) by varying $d_f$. Most surgical lasers produce a Gaussian beam with peak intensity $I_{peak}$, which can be related to $d_f$ through simple optics calculations [7]:

$$d_f = \frac{\pi w^2}{\lambda} \sqrt{\frac{2P}{I_{peak} \pi w^2} - 1} \quad (2)$$

Here, $w$ is the beam waist (i.e., the radius at which the beam intensity fades to $1/e$ of its peak value, measured at the focal point), $\lambda$ is the laser wavelength, and $P$ is the laser optical power.

**Controller Synthesis.** To control the tissue temperature, we synthesize an adaptive controller – this is a well-known family of control methods for systems with uncertain or time-varying parameters. Let us define a control law for the intensity $I_{peak}$ as follows:

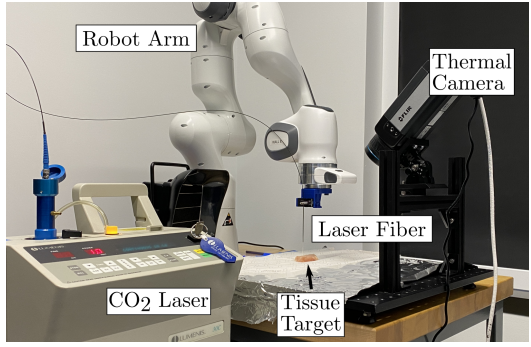

**Fig. 2:** Experimental Setup.

$$I_{peak}(t) = \hat{a}_T T_{peak} + \hat{a}_f f(T_{surf}) + \hat{a}_r r(t) \quad (3)$$

where $t$ represents time, $r(t)$ is a prescribed temperature profile that we wish to create at the laser point of incidence, $T_{surf}$ is the tissue surface temperature, and $T_{peak}$ is the surface temperature at the point of incidence – these latter two quantities are assumed to be measurable with a suitable sensor. In the equation above, $f(T_{surf})$ numerically approximates the heat conduction term in Eq. (1), while $\hat{a}_T$, $\hat{a}_f$, and $\hat{a}_r$ are three scalar coefficients. As sequential temperature measurements become available over time, the adaptive controller updates $\hat{a}_T$, $\hat{a}_f$, and $\hat{a}_r$ in such a way as to minimize the error between the observed temperature $T_{peak}$ and the desired temperature $r$.

**Ex-vivo Experiments.** The controller's performance was verified in experiments on ex-vivo tissue, using the setup shown in Fig. 2. The experiments used a surgical $CO_2$ laser, the Sharplan 30C (Lumenis Ltd., Israel), whose beam is delivered through an optical fiber. The distance $d_f$ between the fiber tip and the tissue is controlled by a Panda robotic arm (Franka Emika GmbH, Germany). The tissue surface temperature is monitored with an A655sc thermal camera (Teledyne FLIR, Oregon, USA) at a rate of 100 fps.

We carried out experiments on four different types of tissue, namely, soft tissue phantoms (2% agar gelatin), bovine liver, bovine bone, and chicken muscle. We note that these tissue types are not representative of what is normally encountered in laryngeal microsurgery and that our goal in this proof-of-principle study is to evaluate the controller's performance for a wide range of optical and thermal tissue properties.

In each experiment, we prescribed a temperature profile $r(t)$ which linearly ramps up to 50 °C, then remains constant for 70 seconds. We carried out five repetitions for each experimental condition, for a total of 20 experiments. All experiments used the same initial values for the controller's parameters, i.e., $\hat{a}_T = 0.152$, $\hat{a}_f = -0.288$, and $\hat{a}_r = 1$, which were obtained by tuning the controller's response on the gelatin phantoms.

## RESULTS

Results are shown in Fig. 3. We observed a tracking error (RMSE) of 2.46 °C on gelatin, 1.95 °C on liver, 1.84 °C on bone, and 2.07 °C on muscle, with standard deviations of 0.2 °C, 0.09 °C, 0.25 °C, and 0.29 °C, respectively.

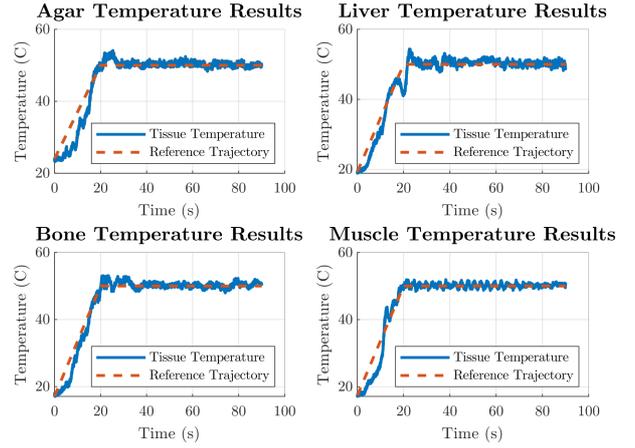

**Fig. 3:** Experimental results. Of the five trials performed on each tissue type, here we report the ones where we observed the median within-group RMSE.

## DISCUSSION

Results indicate the viability of controlling tissue heating by robotically regulating the laser focus. The proposed controller achieved consistent temperature tracking across all four experimental conditions, suggesting that it is robust to variations in the physical properties of the tissue being irradiated. While these results have been obtained with a specific laser ($CO_2$), the proposed controller can be rapidly adapted to work with other types of surgical lasers, so long the wavelength $\lambda$ is known.

An obvious limitation of the proposed controller is that it can only regulate the temperature at a single spot. In future work, we plan to extend the approach presented herein to enable temperature control of a user-defined region of interest. This will require (a) extending the controller to generate 3D motion profiles for the laser and (b) extending the temperature dynamics formulation of Eq. (1) to account for the movement of the laser.